# Incoporating Weighted Board Learning System for Accurate Occupational Pneumoconiosis Staging

Kaiguang Yang, Yeping Wang, Qianhao Luo, Xin Liu, Weiling Li

*Abstract*—Occupational pneumoconiosis (OP) staging is a vital task concerning the lung healthy of a subject. The staging result of a patient is depended on the staging standard and his chest X-ray. It is essentially an image classification task. However, the distribution of OP data is commonly imbalanced, which largely reduces the effect of classification models which are proposed under the assumption that data follow a balanced distribution and causes inaccurate staging results. To achieve accurate OP staging, we proposed an OP staging model who is able to handle imbalance data in this work. The proposed model adopts gray level co-occurrence matrix (GLCM) to extract texture feature of chest X-ray and implements classification with a weighted broad learning system (WBLS). Empirical studies on six data cases provided by a hospital indicate that proposed model can perform better OP staging than state-of-the-art classifiers with imbalanced data.

*Keywords*—*Occupational pneumoconiosis staging*, *Imbalanced Data*, *Weighted broad learning system*, *gray level co-occurrence matrix*.

## I. Introduction

Occupational pneumoconiosis (OP) staging is a vital task concerning the lung healthy of a subject. The level of OP depends on standards provided by authoritative organizations like International Labour Organization (ILO) and National Health Commission of People's republic of China (NHC). According to mainstream standards, e.g., GBZ70-2015, digital chest X-rays are the main basis for OP diagnosis and staging. Hence, OP staging is essentially an image classification.

With the rapid development of artificial intelligence [1-5], Computer-aided diagnosis (CAD) has been widely used in the field of medical imaging classification and many studies concern on classification, especially the chest X-ray classification are proposed [6-8]. Morishita *et al.* first introduced computer-aided diagnosis into pneumoconiosis detection, which was carried out on chest X-ray, they adopted Fourier transformation to obtain the spectrum of lung texture to classify the abnormal areas of lung fields. Murray *et al.*[9] utilized frequency modulation and amplitude modulation analysis for chest X-ray analysis. With the analysis results, it can detect and classify the opaque part of X-ray image according to ILO OP classification standard. Okumura *et al.*[10] used power spectrum to analyze the frequency of OP tablets. They used a rule-based neural network as their classifier for OP staging. In their latter works [11, 12], they enhanced the image with window function and top hat transformation and utilized gray level co-occurrence matrix for better feature representation. To achieve better classification accuracy, they proposed a three-layer neural network.

However, most existing OP staging models take full chest X-ray images and their corresponding labels as inputs. According to mainstream OP staging standards, a final label is an integration of six sub-labels (details can be found in section II). With such inputs, the bottleneck of these models in classification accuracy is obvious. Hence, instead of classifying the full chest X-rays, an effective and practical model for OP staging needs stage each sub-region of lung separately. However, according to an OP data set collected by an occupational disease prevention and control hospital in Chongqing, China from 2019 to 2021, the labels of sub-region of lung are extreme imbalance.

Note that OP staging is essentially an image classification task and most of the classification models are proposed under the assumption that data follow a balanced distribution. When data imbalance occurs, the model whose target is the overall classification accuracy will pay much attention to samples in the majority class, causing the OP staging results to be inaccurate. Data-oriented methods, e.g., over-sampling method and under-sampling method, and model-oriented methods, e.g., cost-sensitive learning, are proposed to address this issue. However, since the size of an OP data set is limited by privacy policy, human resource cost and other constraints, using a data-oriented method or a model-oriented method alone is difficult to solve the data imbalance problem while maximizing the use of known data. To reduce inaccurate OP staging results caused by imbalance data, fusing data-oriented or model-oriented methods with an ensemble learning framework is a possible solution.

Motivated by the above questions and research progress, this study firstly proposes a model for OP staging with imbalanced data. The proposed model takes a set of chest X-ray of a lung sub-region and their labels as input. By calculating the gray level co-occurrence matrix, it obtains feature vectors of each X-ray. Note that the labels corresponding to the samples are extremely imbalanced, to achieve accurate staging results, a weighted broad learning system, which is inspired by [13], is designed to implement the staging process.

To the authors' best knowledge, this is the first study that concerns the data imbalance, which is a common problem in OP staging, thereby achieving an effective approach to implement accurate OP staging. Empirical studies on six data cases provided and authorized by a hospital indicate that proposed model can perform better OP staging than state-of-the-art classifiers with imbalanced data. The remainder of this paper is organized as follows: Section II gives the background of OP staging; Section III presents the details of proposed model; Section IV states the empirical studies; and finally, Section V concludes this paper.

---

- *Corresponding Author: Weiling Li.*
- *K. Yang, Q. Luo and W. Li are with the School of Computer Science and Technology, Dongguan University of Technology, Dongguan, Guangdong, China.*
- *Y. Wang, X. Liu are with the Department of Occupational Disease, Chongqing prevention and treatment center for occupational diseases, Chongqing, China*

## II. BACKGROUND OF OP STAGING

To achieve standard and consistent OP staging results, authoritative organizations like International Labor Organization (ILO) and National Health Commission (NHC) of China have made standards for OP staging. Note that the stage of OP of a lung sub-region is labeled by doctors according to the standards, the data must comply with the same standard.

In this work, the OP data set is provided by an occupational disease prevention and control hospital in China, which abides by last standard authorized by National Health Commission of China [14, 15], i.e., GBZ70-2015. GBZ70-2015 provides a standardized operation process, which can be divided into four steps:

Step 1. Separating the lung area into 6 sub-regions as shown in fig. 1.

Step 2. Evaluating the level of profusion of opacities on each sub-region independently with multiple OP diagnostic physicians and labeling each sub-region with different level, e.g., Normal, Stage 1, Stage 2, and Stage 3.

Step 3. Voting based on each doctor's diagnosis to determine its final level for each sub-region.

Step 4. According to table I, the final OP staging result can be obtained by analyzing all six sub-regions of the chest X-ray.

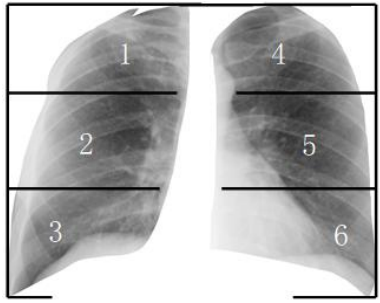

Figure 1. Example of OP Chest X-ray Segmentation.

According to the standardized operation process, a classifier can focus on classifying the level of sub-regions hence to achieve the final staging result by logical operation based on table I.

TABLE I: RULE-BASED DETERMINATION OF FINAL STAGE ACCORDING TO GBZ70-2015.

| STAGE | DESCRIPTION |
| --- | --- |
| Normal | No opacities discover, or Level 1 profusion of opacities presented in one sub-region |
| Stage I | Level 1 profusion of opacities presented in more than two sub-regions, or Level 2 profusion of opacities presented in four sub-regions or less |
| Stage II | Level 2 profusion of opacities presented in four sub-regions or more, or Level 3 profusion of opacities presented |
| Stage III | Large opacities presented |

## III. PROPOSED OP STAGING MODEL

In this section, we proposed an OP staging model with two components for implementing feature extraction [16-21] and classification.

### A. Gray Level Co-occurrence Matrix

Gray level co-occurrence matrix (GLCM) [22] is an effective method to obtain features of chest X-rays since GLCM can extract the lung texture information from chest X-ray effectively.

A chest X-ray is a gray level matrix. A GLCM, e.g., $U$, is a $N \times N$ matrix, where $N$ is the amount of gray levels. Hence, a GLCM is a table to record the number of specific pair of elements in a gray level matrix. Parameters $dx$ and $dy$ are used to determine the spatial relation of two elements to be counted for a GLCM. For example, when $dx=1$ and $dy=0$ as shown in Fig. 2, the value of element $U_{0,1}$ in $U$ is the amount 0-1 element pairs in the gray level matrix whose horizontal distance is 1. Hence, a gray level matrix has multiple GLCMs with different offset parameters.

(a) A Gray Level Matrix  (b) GLCM corresponding to (a)
Figure 2. GLCM of offset parameters $dx=1$ and $dy=0$.

In order to extract feature vector from chest X-ray, we need to calculate four important statistics characteristics [23] from GLCM, e.g., *Energy*, *Entropy*, *Inverse variance* and *contrast*.

- **Energy:** It determines the uniformity of image gray distribution and texture thickness, it can be derived as follows:

$$Eng = \sum_{i}^{N}\sum_{j}^{N} h(i, j)^2, \tag{1}$$

where $h(i, j)$ is the value of the element $i, j$ in GLCM, $N$ is the dimension of GLCM. If the element values of the GLCM are similar, the energy is small, indicating that the texture is consistent, or otherwise.

- **Entropy**: It measures the randomness of an image's information, it can be derived as follows:

$$Ent = -\sum_{i}^{N}\sum_{j}^{N} h(i, j)^2 \cdot \log h(i, j), \tag{2}$$

It indicates the complexity of image gray level distribution.

- **Inverse variance**: It determines the clarity and regularity of texture, which can be obtained as follows:

$$Idm = \sum_{i}^{N}\sum_{j}^{N} h(i,j)/\left(1+(i-j)^2\right), \tag{3}$$

A larger *Idm* indicates clearer texture.

- **Contrast**: It presents the level of partial diversity of an image, which can be obtained as follows:

$$Con = \sum_{i}^{N}\sum_{j}^{N}(i-j)^2 h(i,j)^2. \tag{4}$$

While the level of partial diversity is higher, the contrast is larger.

TABLE II OFFSET PARAMETER SETTING

| No. | Offset parameter |
|---|---|
| $k=1$ | $dx=1, dy=0$ |
| $k=2$ | $dx=0, dy=1$ |
| $k=3$ | $dx=2, dy=0$ |
| $k=4$ | $dx=1, dy=1$ |

In this work, four different offset parameters of GLCM are utilized as shown in table II. Hence, four groups of features can be obtained from different GLCMs. Let $\tau_k$ denote the feature vector of GLCM with $k$-th group of offset parameter, it can be obtained as follows:

$$\tau_k = (Eng_k, Con_k, -Ent_k, Idm_k), \tag{5}$$

where $Eng_k$ denotes the **Energy** of $k$-th group of offset parameter, $Con_i$ denotes the **Contrast** of $k$-th group of offset parameter, $Ent_k$ is **Entropy** of $k$-th group of offset parameter, $Idm_k$ is **Inverse variance** of $k$-th group of offset parameter.

By combining the feature vectors, the input of the classifier, e.g., $\rho$, can be obtained as follows:

$$\rho = (\tau_1, \tau_2, \tau_3, \tau_4) \tag{6}$$

*B. OP Staging with Weighted Broad Learning System*

In this section, a classifier inspired by Weighted Broad Learning System (WBLS) [11] is proposed to implementing accurate OP staging with feature vectors obtained by using GLCM.

To solve the label imbalanced problem, WBLS introduces a weight matrix $C_w$ to provide different weights to different samples in classification.

Assuming we have an input $X=\{(\rho_1, l_1), (\rho_2, l_2),\ldots, (\rho_n, l_n)\}$ with $n$ samples, where $\rho_i$ and $l_i$ denote the feature vector and the label of the $i$-th sample. The ground truth label $l=\{l_1, l_2,\ldots, l_n\}$ is a $n \times m$ matrix, where $n$ is the number of one hot vectors and $m$ the number of staging categories. As shown in Fig. 3, we randomly generate weight matrices $\{W_{f1}, W_{f2},\ldots, W_{fp}\}$ and bias matrices $\{\beta_{f1}, \beta_{f2},\ldots, \beta_{fp}\}$, then map the input $X$ into $i$-th feature node, e.g., $Z_i$, as follows:

$$Z_i = \varphi(XW_{fi} + \beta_{fi}), \tag{7}$$

where $\varphi$ denotes the transformation function for feature layer and $p$ the number of feature nodes. The feature layer, e.g., $Z$, can be obtained by combining feature nodes $(Z_1, Z_2,\ldots, Z_p)$. Similar, we randomly generate weights matrices $\{W_{e1}, W_{e2},\ldots, W_{eq}\}$ and biases matrices $\{\beta_{e1}, \beta_{e2},\ldots, \beta_{eq}\}$, and map $Z_r$ to an enhancement node, e.g., $H_s$, as follows:

$$H_j = \zeta(XW_{ej} + \beta_{ej}), \tag{8}$$

where $\zeta$ denotes the transformation function for enhancement layer and $q$ the number of enhancement nodes. The enhancement layer, e.g., $H$, can be obtained by combining enhancement nodes $(H_1, H_2,\ldots, H_q)$. With $Z$ and $H$, the hidden layer, e.g., $A$, can be easily generated.

The key to use the imbalanced OP data for accurate OP staging with WBLS is to find a weight matrix $W$. It is an minimization problem whose objective is as follows:

$$\arg\min_{W} : C_w \|AW - l\|_2^2 + \lambda \|W\|_2^2, \tag{9}$$

where $C_w$ is a $N \times N$ diagonal matrix used to address imbalanced problem and an element in $C_w$, e.g., $C_{ii}$, can be obtained as follows:

$$C_{ii} = 1/N_{ik}, \quad (10)$$

where $N_{ik}$ denotes the total number of samples contained in category $k$.

Instead of solving (9) with optimization method [24-32], $W$ can be derived by using pseudo inverse method as follows:

$$W = \left(\lambda I + A^T C_w A\right)^{-1} A^T C_w l. \quad (11)$$

With (7), (8) and (11), a classifier for utilizing imbalanced data can be generated.

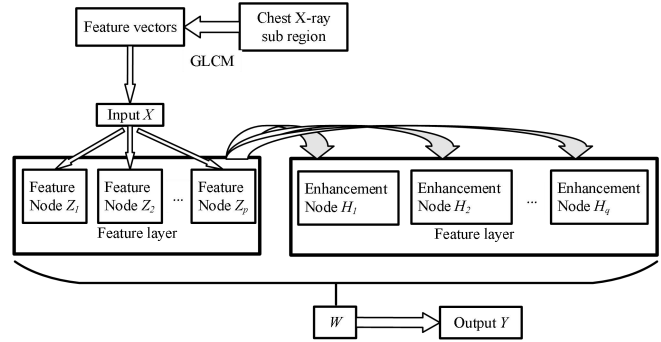

Figure.3 Overview of proposed model

## IV. Experiment

In this section, experiments on real OP data are conducted to prove that our proposed model outperforms models which are built with state-of-the-art feature extraction methods and classifiers in OP staging accuracy.

**Evaluation Metric.** To evaluate the performance of classification, we adopt Accuracy as the evaluation metric[15, 33].

$$\text{Accuracy} = \frac{TN + TP}{FP + FN + TP + TN}, \quad (12)$$

where $TP$ denotes the number of true positive cases in results, $FP$ the number of false positive cases, $FN$ the number of false negative cases, and $TN$ the number of true negative cases.

**Data case.** A data set consists of 678 chest X-ray sub-regions provided by an occupational disease prevention and control hospital in Chongqing, China, is used to conduct the experiment. The data set is divided into six data cases, e.g., C1-C6 according to their sub-region name, namely left-top, left-middle, left-bottom, right-top, right-middle and right bottom.

In our experiments, 75% of samples are selected randomly from the data case to train the model, and the remained as test set with a balancing process, e.g., we ensure that each OP stage has the same number of samples by removing samples from majority randomly. We repeat each group of experiment for 10 times for fair results.

**Optimal algorithm.** The nodes of feature layer and enhancement layer are set to 10 in this case. We use tanh as the transformation function for feature layer and for enhancement layer, its transformation function is sigmoid.

**Compare models.** Six models are involved in our experiments, their details are listed below:

TABLE III: INVOLVED MODELS

| No. | Description |
|---|---|
| M1 | The OP staging model proposed in this work. |
| M2 | An OP staging model which employs singular value decomposition (SVD) as its feature extraction method and implements staging with WBLS. |
| M3 | An OP staging model which employs principle component analysis (PCA) as its feature extraction method and implements staging with WBLS. |
| M4 | An OP staging model which employs GLCM as its feature extraction method and implements staging with Support Vector Machine (SVM). |
| M5 | An OP staging model which employs SVD as its feature extraction method and implements staging with SVM. |
| M6 | An OP staging model which employs PCA as its feature extraction method and implements staging with SVM. |

Experiments results are shown in table. IV. From them, we can see that M1 achieves better OP staging accuracy than its peers. On C1, M1's accuracy is 0.73, 10.6% higher than accuracy at 0.66 by M2, 12.3% higher than accuracy at 0.65 by M3, 40.4% higher than accuracy at 0.52 by M4, 40.4% higher than accuracy at 0.52 by M5, 40.4% higher than accuracy at 0.52 by M6. On C5, M1's accuracy is 0.75, 17.2% higher than accuracy at 0.64 by M2, 10.2% higher than accuracy at 0.68 by M3, 44.0% higher than accuracy at 0.42 by M4, 47.2% higher than accuracy at 0.39 by M5, 38.6% higher than accuracy at 0.45 by M6.

TABLE IV: AVERAGE ACCURACY OF G1

| ACCURACY | C1 | C2 | C3 | C4 | C5 | C6 |
|---|---|---|---|---|---|---|
| M1 | **0.73** | **0.79** | **0.79** | **0.78** | **0.75** | **0.86** |
| M2 | 0.66 | 0.68 | 0.76 | 0.72 | 0.64 | 0.80 |
| M3 | 0.65 | 0.60 | 0.75 | 0.75 | 0.68 | 0.80 |
| M4 | 0.52 | 0.38 | 0.62 | 0.64 | 0.42 | 0.56 |
| M5 | 0.52 | 0.38 | 0.60 | 0.64 | 0.39 | 0.58 |
| M6 | 0.52 | 0.44 | 0.54 | 0.64 | 0.45 | 0.39 |

## V. Conclusion

The staging result of a patient is depended on the staging standard and his chest X-ray. It is essentially an image classification task. However, the distribution of Occupational pneumoconiosis data is commonly imbalanced, which largely reduces the effect of classification models which are proposed under the assumption that data follow a balanced distribution and causes inaccurate staging results. In this work, we proposed an accurate Occupational pneumoconiosis staging model with two components, e.g., extracting texture features with the gray level co-occurrence matrix and implementing classification with a weighted broad learning system. Empirical studies on six data cases provided by a hospital indicate that proposed model can perform better OP staging than state-of-the-art classifiers with imbalanced data.

In our future work, more feature extraction methods [34-37] and deep learning based classifiers will be investigated to further improve OP staging accuracy.

## References

[1] R. N. Bracewell and R. N. Bracewell. "The Fourier transform and its applications, volume 31999." *McGraw-Hill New York*, 1986.

[2] Y. Sato, C. F. Westin, A. Bhalerao. "Tissue classification based on 3D local intensity structures for volume rendering." *IEEE Transactions on visualization and computer graphics*, 2000, 6(2): 160-180.

[3] X. Luo, M. Zhou, S. Li, Y. Xia, Z. You, Q. Zhu, and H. Leung, "An efficient second-order approach to factorize sparse matrices in recommender systems," *IEEE Trans. Ind. Informat.*, vol. 11, no. 4, pp. 946–956, Aug. 2015.

[4] D. Wu and X. Luo, "Robust Latent Factor Analysis for Precise Representation of High-dimensional and Sparse Data," *IEEE/CAA J. Autom. Sinica*, vol.8, no. 4, pp.796-805, Apr 2021.

[5] X. Luo, M. Shang and S. Li, "Efficient Extraction of Non-negative Latent Factors from High-Dimensional and Sparse Matrices in Industrial Applications," *IEEE 16th Int. Conf. Data Mining*, Dec 2016, pp. 311-319.

[6] P. Soliz, J. Ramachandran. "Computer-assisted diagnosis of chest radiographs for pneumoconioses." *Proceedings of SPIE-The International Society for Optical Engineering*, 2001: 667-675.

[7] D. Wu, X. Luo, M. Shang, Y. He, G. Wang, and X. Wu, "A Data-Characteristic-Aware Latent Factor Model for Web Services QoS Prediction," *IEEE Trans. Knowl. Data Eng.*, vol. 34, no. 6, pp. 2525-2538, Jun. 2022.

[8] X. Luo, M. Zhou, S. Li, Z. You, Y. Xia, and Q. Zhu, "A nonnegative latent factor model for large-scale sparse matrices in recommender systems via alternating direction method," *IEEE Trans. Neural Netw. Learn. Syst.*, vol. 27, no. 3, pp. 579–592, Mar. 2016.

[9] V. Murray, M. S. Pattichis, H. Davis. "Multiscale AM-FM analysis of pneumoconiosis x-ray images." *IEEE International Conference on Image Processing (ICIP)*. IEEE, 2009: 4201-4204.

[10] E. Okumura, I. Kawashita, T. Ishida. "Computerized analysis of pneumoconiosis in digital chest radiography: effect of artificial neural network trained with power spectra." *Journal of digital imaging*, 2011, 24(6): 1126-1132.

[11] E. Okumura, I. Kawashita, T. Ishida. "Development of CAD based on ANN analysis of power spectra for pneumoconiosis in chest radiographs: effect of three new enhancement methods." *Radiological physics and technology*, 2014, 7(2): 217-227.

[12] E. Okumura, I. Kawashita, T. Ishida. "Computerized classification of pneumoconiosis on digital chest radiography artificial neural network with three stages." *Journal of digital imaging*, 2017, 30(4): 413-426.

[13] K. Yang, Z. Yu, C. L. P. Chen. "Incremental weighted ensemble broad learning system for imbalanced data." *IEEE Transactions on Knowledge and Data Engineering*, 2021.

[14] A. Krizhevsky. "Convolutional deep belief networks on cifar-10." *Unpublished manuscript*, 2010

[15] Y. Zhang. "Computer-Aided Diagnosis for Pneumoconiosis Staging Based on Multi-scale Feature Mapping." *International Journal of Computational Intelligence Systems*, 2021, 14(1): 1-11.

[16] D. Wu, M. Shang, X. Luo, and Z. Wang, "An $L_1$-and-$L_2$-Norm-Oriented Latent Factor Model for Recommender Systems," *IEEE Tran. Neural Newt. Learn. Syst.*, doi:10.1109/TNNLS.2021.3071392.

[17] X. Luo, Y. Yuan, M. Zhou, Z. Liu, and M. Shang, "Non-Negative Latent Factor Model Based on β-Divergence for Recommender Systems," *IEEE Trans. Syst. Man Cybern., Syst.*, vol. 51. no. 8, pp. 4612-4623, Aug 2021.

[18] M. Shang, Y. Yuan, X. Luo, and M. Zhou, "An α–β-Divergence-Generalized Recommender for Highly Accurate Predictions of Missing User Preferences," *IEEE Trans. Cybern.*, vol. 52, no. 8, pp. 8006-8018, Aug 2021.

[19] X. Luo, M. Zhou, Y. Xia, and Q. Zhu, "An efficient non-negative matrix factorization-based approach to collaborative filtering for recommender systems," *IEEE Trans. Ind. Informat.*, vol. 10 no. 2, pp. 1273–1284, May. 2014.

[20] S. G. Armato III, M. L. Giger, H. MacMahon. "Automated detection of lung nodules in CT scans: preliminary results." *Medical physics*, 2001, 28(8): 1552-1561.

[21] X. Luo, D. Wang, M. Zhou, and H. Yuan, "Latent Factor-Based Recommenders Relying on Extended Stochastic Gradient Descent Algorithms," *IEEE Trans. Syst. Man, Cybern., Syst.*, vol 51, no. 2, pp. 916-926, Feb 2021.

[22] M. Partio, B. Cramariuc, M. Gabbouj. "Rock texture retrieval using gray level co-occurrence matrix." *Nordic Signal Processing Symposium*. 2002, 75.

[23] B. Ginneken, B. M. T. H.Romeny, M. A. Viergever. "Computer-aided diagnosis in chest radiography: a survey." *IEEE Transactions on medical imaging*, 2001, 20(12): 1228-1241.

[24] X. Luo, Y. Zhou, Z. Liu, and M. Zhou, "Fast and Accurate Non-negative Latent Factor Analysis on High-dimensional and Sparse Matrices in Recommender Systems," *IEEE Trans. Knowl. Data Eng.*, doi:10.1109/TKDE.2021.3125252.

[25] X. Luo, Z. Liu, S. Li, M. Shang, and Z. Wang, "A Fast Non-Negative Latent Factor Model Based on Generalized Momentum Method," *IEEE Trans. Syst. Man Cybern., Syst.*, vol 51, no. 1, Jan 2021.

[26] D. Wu, Q. He, M. Shang, Y. He, and G. Wang, "A posterior-neighborhood-regularized latent factor model for highly accurate web service QoS prediction," *IEEE Trans. Serv. Comput.*, vol. 15, no. 2, pp. 793-805, Apr 2022.

[27] X. Luo, H. Wu, Z. Wang, J. Wang, and D. Meng, "A Novel Approach to Large-Scale Dynamically Weighted Directed Network Representation," *IEEE Trans. Pattern Anal. Mach. Intell.*, doi:10.1109/TPAMI.2021.3132503.

[28] K. Awai, K. Murao, A. Ozawa, et al. "Pulmonary nodules at chest CT: effect of computer-aided diagnosis on radiologists detection performance." *Radiology*, 2004, 230(2): 347-352.

[29] X. Luo, H. Wu, H. Yuan, and M. Zhou, "Temporal Pattern-Aware QoS Prediction via Biased Non-Negative Latent Factorization of Tensors," *IEEE Trans. Cybern.*, vol. 50, no.8, May 2020.

[30] Q. Li, S. Sone, K. Doi. "Selective enhancement filters for nodules, vessels, and airway walls in two‐ and three‐ dimensional CT scans." *Medical physics*, 2003, 30(8): 2040-2051.

[31] X. Luo, Y. Zhou, Z. Liu, L. Hu, and M. Zhou, "Generalized nesterov's acceleration-incorporated non-negative and adaptive latent factor analysis," *IEEE Trans. Serv. Comput.*, doi:10.1109/TSC.2021.3069108.

[32] X. Luo, W. Qin, A. Dong, K. Sedraoui, and M. Zhou, "Efficient and High-quality Recommendations via Momentum-incorporated Parallel Stochastic Gradient Descent-Based Learning," *IEEE/CAA J. Autom. Sinica*, vol. 8, no. 2, pp. 402-411, Feb 2021.

[33] S. Binay, P. Arbak, A. A. Safak, et al. "Does periodic lung screening of films meets standards?." *Pakistan Journal of Medical Sciences*, 2016, 32(6): 1506.

[34] X. Luo, M. Zhou, S. Li, Y. Xia, Z. You, Q. Zhu, and H. Leung, "Incorporation of efficient second-order solvers into latent factor models for accurate prediction of missing QoS data," *IEEE Trans.Cybern.*, vol.48, no. 4, pp. 1216-1228, Apr 2018.

[35] S. Mukherjee, S. A. Ahmed, D. P. Dogra. "Fingertip detection and tracking for recognition of air-writing in videos." *Expert Systems with Applications*, 2019, 136: 217-229.

[36] X. Luo, H. Wu, and Z. Li, "NeuLFT: A Novel Approach to Nonlinear Canonical Polyadic Decomposition on High-Dimensional Incomplete Tensors," *IEEE Trans. Knowl. Data Eng.*, doi:10.1109/TKDE.2022.3176466

[37] X. Luo, Y. Yuan, S. Chen, N. Zeng and Z. Wang, "Position-Transitional Particle Swarm Optimization-incorporated Latent Factor Analysis," *IEEE Trans. Knowl. Data Eng.*, vol 34, no. 8, pp. 3958-3970, Aug. 2022.